\documentclass{article}

 \usepackage[main, final]{neurips_2025}
\usepackage{svg}
\usepackage{subcaption}
\usepackage{array}
\workshoptitle{Foundation Models for the Brain and Body}

\usepackage[utf8]{inputenc} 
\usepackage[T1]{fontenc}    
\usepackage{hyperref}       
\usepackage{url}            
\usepackage{booktabs}       
\usepackage{amsfonts}       
\usepackage{nicefrac}       
\usepackage{microtype}      
\usepackage{xcolor}         
\usepackage{amsmath}
\title{Adapting Neural Audio Codecs to EEG}

\author{%
  Ard Kastrati\\
  ETH Zurich\\
  \texttt{akastrati@ethz.ch} \\
  \And
  Luca Lanzend\"{o}rfer \\
  ETH Zurich \\
  lanzendoerfer@ethz.ch
 \\
  \AND
  Riccardo Rigoni \\
  ETH Zurich \\
  rrigoni@ethz.ch \\
  \And
  John Staib Matilla \\
  ETH Zurich \\
  sjohn@ethz.ch \\
  \And
  Roger Wattenhofer \\
  ETH Zurich \\
  wattenhofer@ethz.ch \\
}

\begin{document}

\maketitle

\begin{abstract}
EEG and audio are inherently distinct modalities, differing in sampling rate, channel structure, and scale. Yet, we show that pretrained neural audio codecs can serve as effective starting points for EEG compression, provided that the data are preprocessed to be suitable to the codec’s input constraints. Using DAC, a state-of-the-art neural audio codec as our base, we demonstrate that raw EEG can be mapped into the codec’s stride-based framing, enabling direct reuse of the audio-pretrained encoder-decoder. Even without modification, this setup yields stable EEG reconstructions, and fine-tuning on EEG data further improves fidelity and generalization compared to training from scratch. We systematically explore compression-quality trade-offs by varying residual codebook depth, codebook (vocabulary) size, and input sampling rate. To capture spatial dependencies across electrodes, we propose DAC-MC, a multi-channel extension with attention-based cross-channel aggregation and channel-specific decoding, while retaining the audio-pretrained initialization. Evaluations on the TUH Abnormal and Epilepsy datasets show that the adapted codecs preserve clinically relevant information, as reflected in spectrogram-based reconstruction loss and downstream classification accuracy.
\end{abstract}


\section{Introduction and Related Work}

Electroencephalography (EEG) plays a central role in clinical neurology and neuroscience, enabling the non-invasive monitoring of brain activity in applications such as epilepsy diagnosis, sleep staging, and cognitive assessment. As machine learning becomes increasingly integrated into healthcare~\cite{medpalm2, healthbench}, there is growing interest in building large-scale models that can generalize across EEG datasets, subjects, and clinical conditions. Inspired by the success of foundation models in computer vision~\cite{awais2023foundational}, language~\cite{wei2022emergent} and audio~\cite{huang2023masked}, early efforts are now being made to pretrain models on EEG at scale as well~\cite{bendr, wu2022neuro2vec, cui2024neurogpt, jiang2024large}. A key enabler of such approaches is the ability to represent EEG signals compactly and discretely, making them easier to store, index, and model using architectures originally designed for token sequences. Neural codecs have recently shown promise in this direction for other modalities, particularly audio, where they compress raw signals into discrete tokens while preserving high reconstruction fidelity. This enables sequence models to process continuous signals as token streams, facilitating next-token prediction in time-series data using the same modeling strategies that have proven effective in natural language.

Recent advances in neural compression techniques~\cite{zeghidour2021soundstream, kumar2023high, encodec, siuzdak2024snac, lanzendorfer2024neural} have revolutionized audio compression by employing residual vector quantization within adversarially trained autoencoders. These methods are usually trained with a large corpora of audio datasets. However, EEG is far less common as a data resource, as its collection is expensive, time-consuming, and subject to strict privacy and ethical constraints. As a result, publicly available EEG datasets are orders of magnitude smaller than those for audio, making it considerably harder to train general-purpose codecs directly on EEG. With limited data, codecs struggle not only to generalize across different recording setups and subject populations, but even to avoid overfitting within a given dataset.

Despite the substantial differences between audio and EEG in sampling rates, channel structure, and signal characteristics, we find that pretrained audio codecs can serve as an unexpectedly strong starting point for EEG compression. In fact, we show that simply feeding EEG signals into an off-the-shelf audio codec already yields surprisingly reasonable reconstructions. Building on this insight, we fine-tune the neural audio codec DAC~\cite{kumar2023high} on EEG recordings and show that this improves reconstruction fidelity and generalization compared to training a codec from scratch. To explore how DAC can be adapted to EEG, we vary key parameters that influence the compression rate, including the number of residual codebooks, the size of the vocabulary, and the internal sampling rate used to present EEG to the codec. In addition, we propose DAC-MC, an extension designed to handle the multi-channel nature of EEG. DAC-MC incorporates attention-based aggregation across channels and channel-specific conditioning in the decoder, enabling compression and reconstruction of multi-channel EEG within a unified framework. Importantly, all variants including the multi-channel extension, are initialized from the pretrained DAC checkpoint trained on 44.1 kHz audio, preserving the benefits of large-scale audio training throughout. 

We evaluate our models on two largest EEG labeled datasets: Abnormal (TUAB) and Epilepsy (TUEP) dataset from the TUH EEG Corpus~\cite{tuh2017}. We use two types of evaluation: (i) reconstruction fidelity, measured using a spectrogram-based loss between the original and reconstructed signals, and (ii) downstream performance, where classification models are applied to reconstructed EEG to assess whether relevant information is preserved. While some performance degradation is observed after reconstruction, the results show that the trained codec retains the essential structure needed for meaningful downstream tasks in the clinical domain.



\section{Methods}

\subsection{From Audio Codec to EEG Codec}

We use the Improved RVQGAN (DAC)~\cite{kumar2023high}, a residual vector-quantized encoder–decoder. Residual quantization builds a stack of codebooks: shallow codebooks capture coarse structure; deeper codebooks add detail. We refer to \cite{kumar2023high} for details and focus here on how we can use this model for EEG compression.

\paragraph{Dealing with sampling rate.} Our main idea behind adapting an audio codec at 44.1\,kHz to EEG at 512\,Hz is simple: we directly feed raw EEG segments (clipped to ±200$\mu$V and normalized to the [-1, 1] range similar to audio scale) into the pretrained model, even though the \textit{temporal meaning} of each segment differs. In DAC~\cite{kumar2023high}, a fixed window determines how many input samples correspond to one output token, i.e., 512 samples per token. While 512 samples corresponds to $\sim$13\,ms in 44.1\,kHz audio, it spans 1\,s at 512\,Hz EEG. Nevertheless, we observe that the pretrained model can still produce stable and interpretable outputs, allowing us to treat EEG compression as a repurposing of an off-the-shelf audio codec.

\begin{figure}[htbp]
  \centering
  \includegraphics[width=\linewidth]{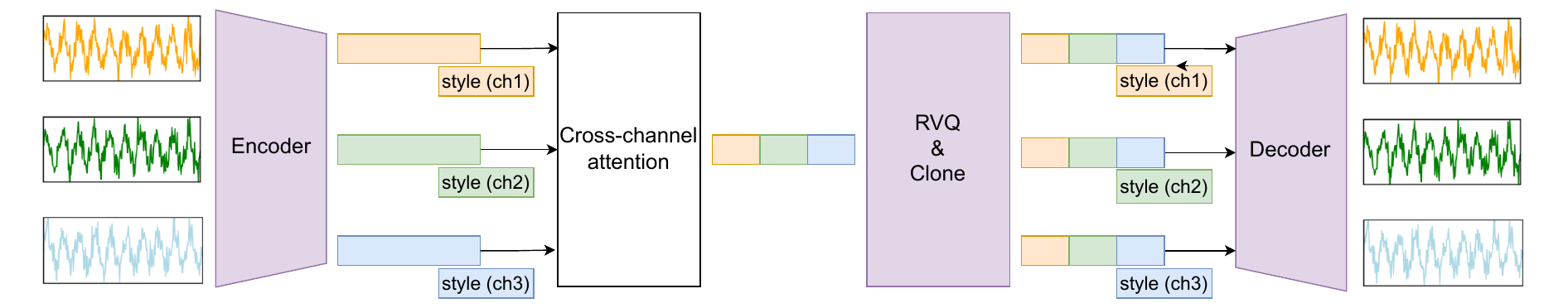}
  \caption{DAC-MC. Purple modules form the pretrained DAC backbone.}
  \label{fig:dac_mc_architecture}
\end{figure}

\paragraph{Dealing with multi channels.}
The simplest way to apply DAC to EEG is to compress each channel independently, using the original audio-pretrained encoder and decoder without modification. We refer to this approach as \texttt{DAC-SC} (DAC Single-Channel). While straightforward, it ignores spatial correlations across electrodes and treats each channel as an isolated 1D signal. On the other hand, to exploit cross-channel dependencies and further increase compression, we introduce \texttt{DAC-MC}, a multi-channel extension that encodes multiple channels jointly. Each EEG channel is first processed by the \texttt{DAC-SC} encoder to produce a latent sequence. These latents are concatenated across channels (dimension: \texttt{hidden\_dim}$\times$\texttt{num\_channels}), with zero-padding applied when needed to support variable-length inputs. We then apply self-attention over the concatenated latents to capture both temporal and inter-channel structure, followed by a projection back to the original hidden dimension for quantization and decoding. To enable channel-specific decoding, the decoder is modulated by learned style vectors (one per channel) which apply affine transformations (scale and bias) to the latent space, inspired by StyleGAN~\cite{style_gan}. Fixed channel ordering and positional encodings are used to preserve spatial context. Because full attention across many channels can be computationally expensive (due to DAC’s 1024-dimensional latent space), we group EEG channels into smaller subsets (variable size, at most 5) and process each group independently. We evaluate two grouping strategies: \emph{Random Groups}, which vary per training step to improve generalization, and \emph{Manual Groups}, which reflect anatomical or montage-based spatial structures (see Appendix~\ref{appendix:groups} for detailed description). Importantly, \texttt{DAC-MC} adds lightweight adapters around the pretrained DAC model, so all components are compatible with audio-based initialization (see Figure~\ref{fig:dac_mc_architecture}).

\paragraph{Configuration Options.} Several modeling decisions remain flexible and influence the trade-off between compression and reconstruction fidelity. These choices let us use the same model in different ways, without modifying its architecture: \textit{(i) Upsampling.} By feeding the model EEG at its native rate (e.g., 256\,Hz), a token corresponds to a longer time span, resulting in stronger compression. Upsampling to 512\,Hz produces more tokens per second, effectively using the model with a finer temporal resolution at the cost of a higher bitrate: 90\,bts (resp.~180\,bts) at 512Hz (resp.~256Hz). \textit{(ii) Vocabulary size.} We also can reduce the vocabulary size of each residual codebook (e.g. from 1024 to 512)  by merging similar entries in the codebook during finetuning. This limits the model's expressive capacity, but also reduces bitrate. Importantly, this variant still uses the pretrained weights as a starting point and is adapted to EEG through fine-tuning. \textit{(iii) Residual depth.} We can also reduce the number of residual codebooks (e.g., from 9 to 3). This controls how many tokens are combined to represent each input. Here we distignuish two types: \emph{(i) post fine-tune pruning:} fine-tune the model with all 9 codebooks, but during evaluation reconstruct signals using only the first $k<9$ codebooks,  \emph{(ii) pre fine-tune pruning:} train a separate model for each residual depth (from 3 to 9 codebooks), yielding 7 distinct fine-tuned models.


\subsection{Datasets and Preprocessing}

\textbf{Datasets.} We used EEG recordings from the publically-available dataset: the Temple University Hospital EEG Corpus (TUH EEG)~\cite{tuh2017}. TUH EEG served as the primary dataset for model fine-tuning. Benchmark evaluation used (labeled) TUAB subset (abnormal EEG detection) and TUEP subset (epilepsy detection). 

\textbf{Preprocessing.} Both datasets underwent identical preprocessing: noisy initial segments were removed, empty channels were excluded, signals were resampled to 512 Hz, and a 0.1 Hz high-pass filter was applied. Amplitude values were clipped to ±200$\mu$V, normalized to the [-1, 1] range, and finally segmented into non-overlapping 30-second windows.

\subsection{Training and Optimization}

We considered three approaches for employing \texttt{DAC-SC} to EEG signals, and namely: training the model from scratch using only EEG data (\texttt{Scratch}), use the pre-trained \texttt{DAC-SC} model checkpoint on 44.1\,kHz audio (\texttt{Pre-trained}), and fine-tune the pre-trained checkpoint on EEG data (\texttt{Fine-tuned}). All models were trained using the Adam optimizer with a learning rate of \texttt{1e-5} and $\beta$ parameters set to $(0.8, 0.999)$, using 4 RTX3090 GPUs.

We employed the composite loss function from the original DAC paper~\cite{kumar2023high}, including waveform, multi-scale STFT, Mel-spectrogram, adversarial (GAN), commitment, and codebook losses. To better match EEG characteristics, the Mel-spectrogram loss was replaced with a standard spectrogram loss. The waveform loss was initially weighted more heavily to guide adaptation to EEG, then gradually reduced to prevent smoothing effects that diminished peak amplitudes and biased outputs toward the signal mean. Adversarial losses proved unstable on EEG data, often diverging due to domain mismatch. To mitigate this, we adopted a two-phase training strategy: GAN losses were included with reduced weight in the first phase, then removed entirely once divergence was observed.

\subsection{Evaluation}

Our evaluation framework measures EEG reconstruction quality with a focus on clinical indistinguishability, analogous to perceptual indistinguishability in audio codecs. The goal is to ensure that reconstructed EEG signals preserve key clinical features relevant for neurological diagnosis and classification. We adopt two complementary evaluation strategies: \textit{(i) Reconstruction loss.} We compute spectrogram-based loss between original and reconstructed signals to quantify how well the model preserves spectral structure over time. \textit{(ii) Downstream classification.} We assess whether clinically relevant information is retained in the reconstructed signals by training classifiers on frequency-domain features extracted from reconstructions alone. We use standard models such as Random Forests and Decision Trees, trained with features from the Brainfeatures\footnote{\url{https://github.com/TNTLFreiburg/brainfeatures}} library. Benchmarks include two widely used clinical classification tasks: (i) anomaly detection using the TUH Abnormal subset (TUAB), and (ii) epilepsy diagnosis using the TUH Epilepsy subset (TUEP). 





\section{Results}

\begin{figure}[htbp]
    \centering
    \includegraphics[width=0.99\linewidth]{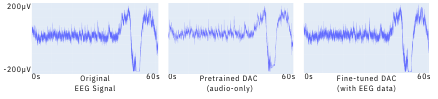}
    \caption{Example reconstruction with audio-pretrained codec and fine-tuned codec with EEG data.}
    \label{fig:single_channel_reconstruction}
\end{figure}

We begin by examining whether audio-pretrained codecs can reconstruct EEG signals. Figure~\ref{fig:single_channel_reconstruction} shows a representative example from the test set of single-channel EEG reconstruction using \texttt{DAC-SC}, where the reconstructed signal closely follows the morphology of the original. In Appendix~\ref{appendix:multichannel} we provide also examples that extend this to multi-channel EEG using \texttt{DAC-MC}, illustrating that spatial and temporal structure is preserved across several electrodes. These examples demonstrate that even though the codec was pretrained on audio, it can produce stable and realistic EEG reconstructions when adapted appropriately. 

\begin{figure*}[htbp]
    \centering
    \begin{subfigure}[t]{0.48\textwidth}
        \centering
        \includegraphics[width=\linewidth]{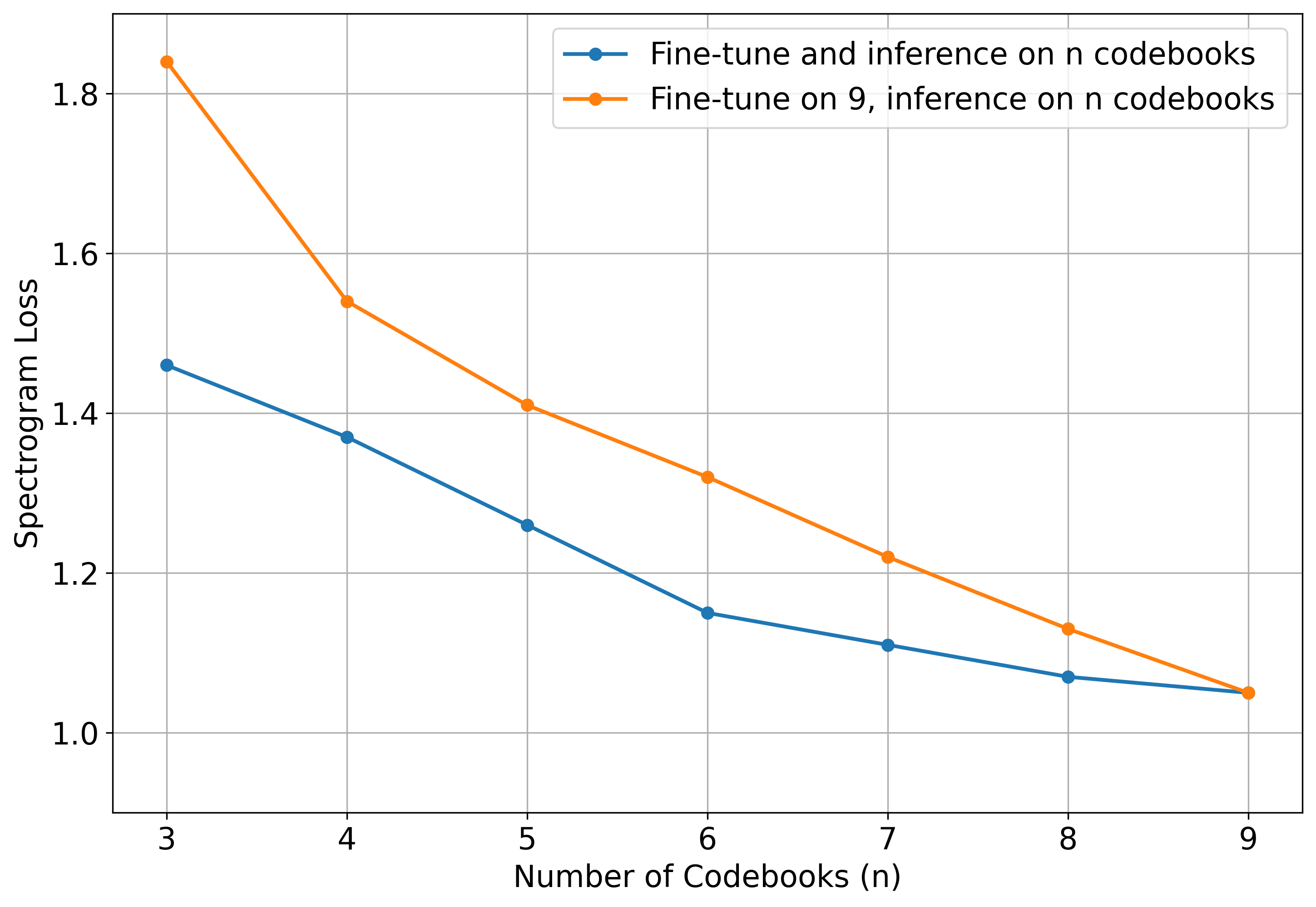}
       \caption{Effect of pruning residual codebooks (last-to-first) from the vector quantizer before and after fine-tuning the \texttt{Audio-to-EEG} model. \emph{After}: the model is fine-tuned with all nine codebooks and, at inference, only the first $n$ are kept. \emph{Before}: the model is fine-tuned from scratch using only the first $n$ codebooks.}

        \label{fig:codebook_analysis}
    \end{subfigure}
    \hfill
    \begin{subfigure}[t]{0.48\textwidth}
        \centering
        \includegraphics[width=\linewidth]{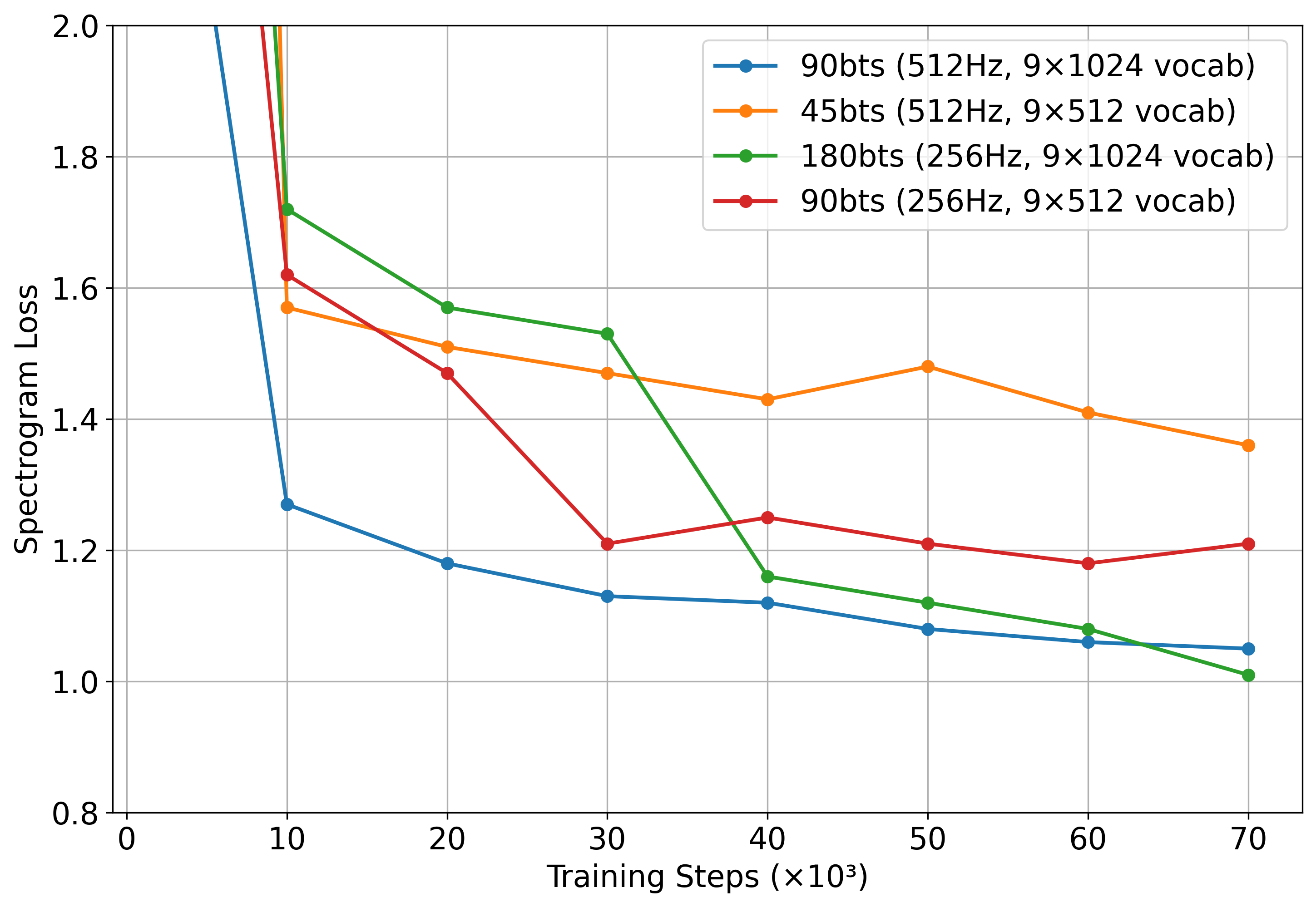}
        \caption{Effect of upsampling and alphabet size reduction on reconstruction fidelity.}
        \label{fig:sampling_rate_analysis}
    \end{subfigure}
    \caption{Comparison of EEG codec adaptations. (a) Losses for different training strategies. (b) Trade-offs with residual codebooks. (c) Sampling rate and alphabet size effects.}
    \label{fig:codec_adaptations}
\end{figure*}

\paragraph{Using pretrained audio codecs is better than training from scratch.}
Table~\ref{tab:pretrained_vs_scratch_summary}  demonstrates the effectiveness of fine-tuning the DAC model, originally pretrained on audio data, for EEG signal reconstruction. The \texttt{Audio-to-EEG Fine-tuned} model achieved significantly lower spectrogram reconstruction loss (approximately 1.05) compared to both the \texttt{Scratch} (1.46) and \texttt{Audio-Pretrained} models without fine-tuning (2.5). Remarkably, the fine-tuned EEG model nearly matched the original DAC model's performance on audio data (1.09, see~\cite{kumar2023high}), highlighting DAC’s robust generalization capabilities.

\paragraph{Impact of codebook size.}
Figure~\ref{fig:codebook_analysis} explores the impact of reducing residual codebooks for higher compression. A reduction from 9 to 6 codebooks resulted in less than a 10\% increase in spectrogram loss, representing a favorable trade-off between computational efficiency and reconstruction accuracy. Conversely, reducing codebooks directly from a fully fine-tuned model consistently produced higher spectrogram losses than fine-tuning with fewer codebooks. Notably, reducing from 9 to 3 codebooks doubled the spectrogram loss for the \texttt{Audio-to-EEG Fine-tuned} model. 

Further hyperparameter experiments, depicted in Figure~\ref{fig:sampling_rate_analysis}, examined the combined effects of increasing the internal sampling rate (upsampling from 256\,Hz to 512\,Hz) and reducing the codebook alphabet size (from 1024 to 512 entries). Upsampling offered marginal improvements in temporal granularity and reconstruction fidelity, which were negligible relative to the substantial increase in computational requirements. Additionally, alphabet size reduction consistently degraded reconstruction quality unless combined with iterative pruning and further fine-tuning.

\paragraph{Downstream Tasks}

\begin{table}[htbp]
  \centering
  \begin{subtable}[t]{0.36\textwidth} 
    \centering
    \begingroup
    \setlength{\tabcolsep}{4pt} 
    \footnotesize
    \renewcommand{\arraystretch}{1.15}
    \begin{tabular}{lc}
      \hline
      \textbf{Model} & \textbf{Loss} \\
      \hline
      \texttt{Audio-Pretrained} & 2.50 \\
      \texttt{Scratch} & 1.46 \\
      \texttt{Audio-to-EEG} & 1.05 \\
      \hline
    \end{tabular}
    \endgroup
    \caption{Comparison of the spectrogram reconstruction losses from test set showing that fine-tuning an audio-pretrained codec for EEG yields the best performance.}
    \label{tab:pretrained_vs_scratch_summary}
  \end{subtable}
  \hfill
  \begin{subtable}[t]{0.62\textwidth} 
    \centering
    \footnotesize
    \renewcommand{\arraystretch}{1.2}
    \begin{tabular}{l|cc}
      \hline
      \textbf{Mode} & \textbf{Epilepsy} & \textbf{Abnormal} \\
      \hline
      Single-Channel (\texttt{DAC-SC}) & 80\,\% & 83\,\% \\
      Single-Channel (\texttt{DAC-MC}) & 82\,\% & 81\,\% \\
      Random-Groups (\texttt{DAC-MC}) & 85\,\% & 78\,\% \\
      Manual-Groups (\texttt{DAC-MC}) & 85\,\% & 78\,\% \\
      \hline
      Baseline & 84\,\% & 82\,\% \\
      \hline
    \end{tabular}
    \caption{Benchmark accuracy for Epilepsy and Abnormal EEG datasets.
  \texttt{DAC-SC} is a single-channel model. \texttt{DAC-MC} is a multi-channel model that can be used either per-channel decoding
  (``Single-Channel (DAC-MC)'') or jointly on multiple channels with groups chosen at random or manually 
  (``Random-Groups''/``Manual-Groups'' (DAC-MC)). The Baseline is the best accuracy obtained when training and testing on the original signals.}
    \label{tab:combined_performance_comparison}
  \end{subtable}
\end{table}


Table~\ref{tab:combined_performance_comparison} reports classification accuracies for epilepsy and abnormal EEG detection using \texttt{DAC-SC} and \texttt{DAC-MC} under different decoding setups. For epilepsy, grouped \texttt{DAC-MC} (Random/Manual Groups) reaches 85\% vs.\ 80\% for \texttt{DAC-SC}, indicating gains from modeling spatial dependencies across channels. For abnormal EEG, however, grouped \texttt{DAC-MC} underperforms at 78\%, while running the same multi-channel model in a \emph{single-channel} mode (i.e., decoding one channel at a time) attains 81\%—closer to the baseline trained and tested on original signals (82\%) and to \texttt{DAC-SC} (83\%). This suggests that, for abnormality detection, per-channel decoding preserves the relevant features better than cross-channel grouping.

\section{Conclusions}
We show that pretrained neural audio codecs can be repurposed for EEG compression with minimal modifications. By fine-tuning DAC on EEG data and introducing a multi-channel extension (DAC-MC), we achieve high-fidelity reconstructions that preserve clinically relevant structure across datasets. Our analysis spans compression-quality trade-offs via sampling rate, codebook size, and depth. 
\clearpage

\bibliographystyle{unsrt}
\bibliography{example_paper}

\appendix


\newpage
\section{Multi-channel grouping}
\label{appendix:groups}

\subsection{Random grouping}
\label{appendix:random_groups}
\begin{figure}[htbp]
  \centering
  \includegraphics[width=\linewidth]{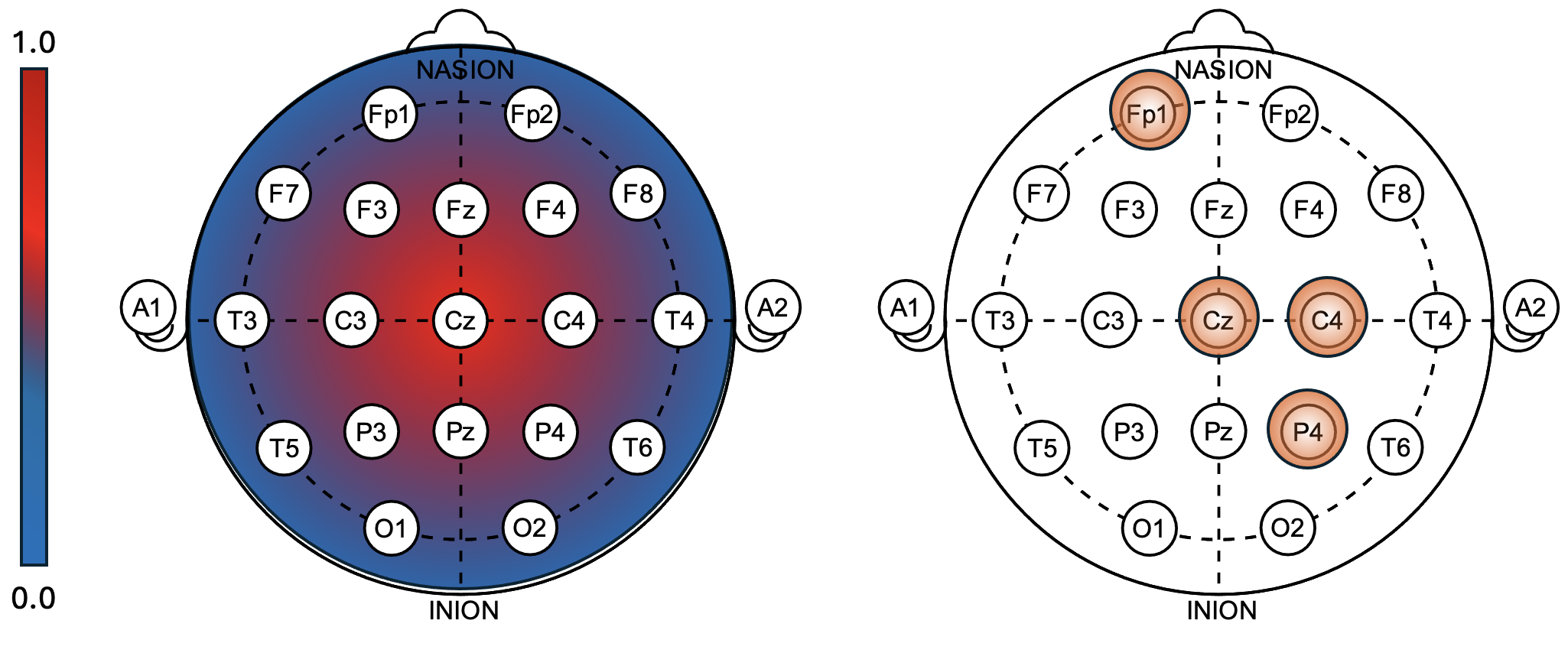}
  \caption{Random subset of 4 channels with pivot Cz on the 10-20 System. On the left, the induced probability distribution from pivot Cz. On the right, the sampled channels.. \emph{Adapted from} \cite{wikipedia_1020_2010}. Public domain.}
  \label{fig:myplot}
\end{figure}

We outline here the \emph{random groups} sampling procedure employed during fine-tuning. For a given EEG recording, we start with all EEG channels $\mathcal{C}=\{1,\dots,N\}$, each with a 3D scalp location
$\mathbf{x}_c \in \mathbb{R}^3$ on the unit sphere ($\|\mathbf{x}_c\|_2=1$). We keep a pool
$\mathcal{R}$ of channels not yet assigned, and build groups $G_1, G_2, \dots$ until
$\mathcal{R}$ is empty.

\paragraph{1) Pick a group size (1–5).}
For each group, draw $X \sim \mathrm{Exp}(\lambda=3)$, round up, and clip to the range $[1,5]$:
\[
s \;=\; \min\{5,\,\max\{1,\,\lceil X\rceil\}\}.
\]
Finally, make sure we don’t ask for more channels than remain: $s \leftarrow \min\{s,\;|\mathcal{R}|\}$.
(If you want larger or smaller typical groups, scale $X$ before rounding.)

\subsection{Manual grouping}
\label{appendix:manual_groups}
We outline here the \emph{manual groups} crafted for the multi-channel EEG reconstruction for the epilepsy and anomaly detection tasks. 


\begin{table}[htbp]
    \centering
    \footnotesize
    \renewcommand{\arraystretch}{1.15}
    \begin{tabular}{>{\raggedleft\arraybackslash}p{1.2cm} | p{0.78\linewidth}}
        \hline
        \textbf{Group} & \textbf{Channels} \\
        \hline
        1 & \texttt{F3}, \texttt{F4}, \texttt{F7}, \texttt{F8} \\
        2 & \texttt{FP1}, \texttt{FP2}, \texttt{P3}, \texttt{P4} \\
        3 & \texttt{T3}, \texttt{T4}, \texttt{T5}, \texttt{T6} \\
        4 & \texttt{C3}, \texttt{C4}, \texttt{CZ} \\
        5 & \texttt{O1}, \texttt{O2} \\
        \hline
    \end{tabular}
    \caption{Manual channel groups for the epilepsy task.}
    \label{tab:epilepsy_manual_groups}
\end{table}

\begin{table}[htbp]
    \centering
    \footnotesize
    \renewcommand{\arraystretch}{1.15}
    \begin{tabular}{>{\raggedleft\arraybackslash}p{1.2cm} | p{0.78\linewidth}}
        \hline
        \textbf{Group} & \textbf{Channels} \\
        \hline
        1  & \texttt{EEG 26-REF}, \texttt{EEG 27-REF}, \texttt{EEG 28-REF}, \texttt{EEG 29-REF} \\
        2  & \texttt{EEG 30-REF}, \texttt{EEG 31-REF}, \texttt{EEG 32-REF} \\
        3  & \texttt{EEG C3-REF}, \texttt{EEG C3P-REF}, \texttt{EEG C4-REF}, \texttt{EEG C4P-REF}, \texttt{EEG CZ-REF} \\
        4  & \texttt{EEG FP1-REF}, \texttt{EEG F3-REF}, \texttt{EEG F7-REF}, \texttt{EEG FZ-REF} \\
        5  & \texttt{EEG F4-REF}, \texttt{EEG FP2-REF}, \texttt{EEG F8-REF} \\
        6  & \texttt{EEG T1-REF}, \texttt{EEG T2-REF}, \texttt{EEG T3-REF}, \texttt{EEG T4-REF}, \texttt{EEG T5-REF} \\
        7  & \texttt{EEG O1-REF}, \texttt{EEG O2-REF}, \texttt{EEG OZ-REF}, \texttt{EEG T6-REF} \\
        8  & \texttt{EEG P3-REF}, \texttt{EEG P4-REF}, \texttt{EEG PG1-REF}, \texttt{EEG PG2-REF}, \texttt{EEG PZ-REF} \\
        9  & \texttt{EEG EKG1-REF}, \texttt{EEG LOC-REF}, \texttt{EEG ROC-REF} \\
        10 & \texttt{EEG A1-REF}, \texttt{EEG A2-REF}, \texttt{EEG SP1-REF}, \texttt{EEG SP2-REF} \\
        \hline
    \end{tabular}
    \caption{Manual channel groups for the TUAB (Abnormal EEG) task.}
    \label{tab:tuab_manual_groups}
\end{table}

\paragraph{2) Choose a pivot.}
Pick one pivot channel $p$ uniformly at random from $\mathcal{R}$ and put it in the group~($G \gets \{p\}$).

\paragraph{3) Fill the rest by proximity.}
For each candidate $c \in \mathcal{R}\setminus G$, compute its distance to the pivot
(using chord distance on the unit sphere, which has the property of being monotonic with the geodesic angle):
\[
d(p,c) \;=\; \|\mathbf{x}_p - \mathbf{x}_c\|_2.
\]
Turn distances into sampling weights so nearer channels are more likely:
\[
w_c = \exp\!\left(-\frac{d(p,c)}{\tau}\right), \qquad
\pi(c \mid p) = \frac{w_c}{\sum_{c' \in \mathcal{R}\setminus G} w_{c'}}.
\]
Sample $s-1$ distinct neighbors without replacement from $\mathcal{R}\setminus G$ using $\pi(\cdot \mid p)$,
add them to $G$, remove $G$ from $\mathcal{R}$, and repeat. We set the temperature $\tau = 1$.

\section{DAC-DC Reconstructions}
\label{appendix:multichannel}
Below we give example reconstructions of 5 neighboring channels with DAC-MC. 

\begin{figure}[htbp]
    \centering
    \includegraphics[width=0.99\linewidth]{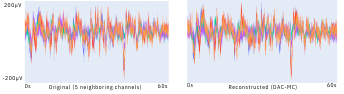}
    \caption{Example reconstruction with fine-tuned codec with multi-channels.}
    \label{fig:multi_channel_reconstruction}
\end{figure}

\end{document}